\documentclass[letterpaper]{article} 
\usepackage{aaai2026}  
\usepackage{times}  
\usepackage{helvet}  
\usepackage{courier}  
\usepackage[hyphens]{url}  
\usepackage{graphicx} 
\usepackage{natbib}  
\usepackage{caption} 
\usepackage{newfloat}
\usepackage{xcolor}
\usepackage{amsmath, amssymb}
\usepackage{extarrows}
\usepackage{multirow} 
\usepackage{graphicx}  
\usepackage{subcaption}

\usepackage[ruled, linesnumbered, noend]{algorithm2e}

\usepackage{colortbl}
\newcommand{\graycell}{\cellcolor{gray!15}}
\usepackage{bbding} 
\usepackage{booktabs} 
\usepackage{arydshln}
\usepackage{utfsym}

\usepackage{hyperref} 
\usepackage{cleveref}
\hypersetup{
    colorlinks=true,   
    linkcolor=cyan,     
    urlcolor=cyan,      
    citecolor=black,    
}

\newcommand{\w}{\mathbf{w}}

\newcommand{\x}{\mathbf{x}}

\def\b{\mathbf{b}}
\def\l{\mathbf{l}}
\newcommand{\J}{\mathbf{J}}
\def\H{\mathbf{H}}

\def\I{\mathbf{I}}

\def\C{\mathbf{C}}
\def\D{\mathbf{D}}
\def\I{\mathbf{I}}
\newcommand{\tW}{{\pmb{\mathcal{W}}}}

\newcommand{\tX}{{\pmb{\mathcal{X}}}}

\newcommand{\tC}{{\pmb{\mathcal{C}}}}
\newcommand{\tD}{{\pmb{\mathcal{D}}}}
\newcommand{\mR}{{\mathbb{R}}}
\def\eg{{e.g.}}
\def\ie{{i.e.}}

\crefname{table}{Table}{Tables}
\crefname{figure}{Figure}{Figures}
\crefname{equation}{Eq.}{Eqs.}
\crefname{algorithm}{Algorithm}{Algorithms}

\title{Optimal Brain Connection: Towards Efficient Structural Pruning}
\author{
   Shaowu Chen\textsuperscript{\rm 1,2},
   Wei Ma\textsuperscript{\rm 1},
   Binhua Huang\textsuperscript{\rm 2},
   Qingyuan Wang\textsuperscript{\rm 2},
   Guoxin Wang\textsuperscript{\rm 2},\\
   Weize Sun\textsuperscript{\rm 1},
   Lei Huang\textsuperscript{\rm 1},
   Deepu John\textsuperscript{\rm 2}
}
\affiliations {
    \textsuperscript{\rm 1}State Key Laboratory of Radio Frequency Heterogeneous Integration, Shenzhen University, China\\
    \textsuperscript{\rm 2}School of Electrical and Electronic Engineering, University College Dublin, Ireland\\
}

\usepackage{bibentry}

\nocopyright
\begin{document}

\maketitle

\begin{abstract}
Structural pruning has been widely studied for its effectiveness in compressing neural networks. However, existing methods often neglect the interconnections among parameters. To address this limitation, this paper proposes a structural pruning framework termed \textbf{Optimal Brain Connection}. First, we introduce the \textbf{Jacobian Criterion}, a first-order metric for evaluating the saliency of structural parameters. Unlike existing first-order methods that assess parameters in isolation, our criterion explicitly captures both intra-component interactions and inter-layer dependencies. Second, we propose the \textbf{Equivalent Pruning} mechanism, which utilizes autoencoders to retain the contributions of all original connections—including pruned ones—during fine-tuning. Experimental results demonstrate that the Jacobian Criterion outperforms several popular metrics in preserving model performance, while the Equivalent Pruning mechanism effectively mitigates performance degradation after fine-tuning. Code: \url{https://github.com/ShaowuChen/Optimal_Brain_Connection}
\end{abstract}


\section{Introduction}
\label{sec:Introduction}

\begin{figure}[ht]
    \centering
    \includegraphics[width=0.46\textwidth]{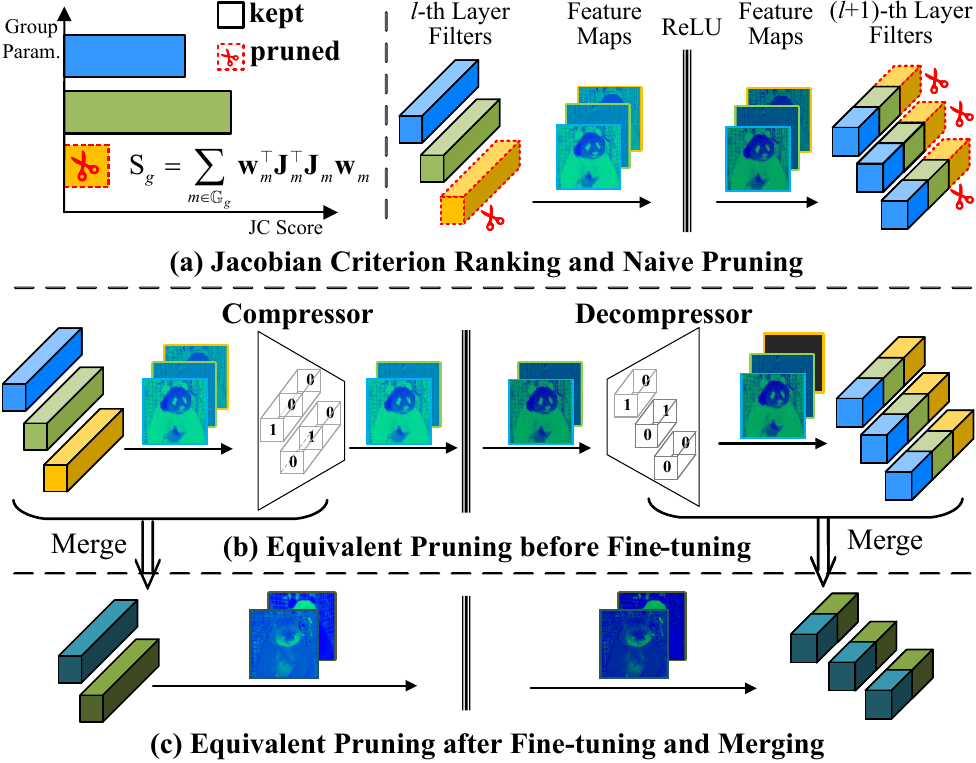}
    \caption{Overview of our OBC framework. (a) Our Jacobian Criterion accurately ranks structural groups by capturing parameter interactions. (b) Our Equivalent Pruning then creates a temporary autoencoder, allowing all connections (including pruned ones) to contribute to parameter recalibration and accuracy recovery during fine-tuning. (c) Finally, the autoencoder is permanently merged after fine-tuning, resulting in an efficient pruned model with the same structure as naive pruning but with better performance.}
    \label{fig:framework}
\end{figure}

Despite the remarkable success of deep neural networks across various domains, their increasing scale poses challenges for deployment on resource-constrained devices. To address this issue, pruning methods—including unstructured and structural pruning—have been developed as effective solutions. Unstructured pruning \cite{gadhikar2024masks,han2015learning} zeros out element-wise parameters, leading to irregular sparsity that necessitates customized software and hardware to accelerate networks. In contrast, structural pruning \cite{fang2023depgraph} removes redundant groups of components (such as filters and channels), resulting in a slimmer network with improved inference efficiency and thus attracting increasing attention in recent years.

A core task of structural pruning involves identifying redundant components that can be removed without severely degrading network performance. To this end, a variety of data-free criteria have been proposed, including norm-based~\cite{fang2024isomorphic}, relationship-based~\cite{LRF2021Joo}, and hybrid criteria~\cite{chen2023whc}. In contrast, data-driven criteria evaluate the ``saliency'' of parameters~\cite{IFSO}, \ie, the change in the loss function induced by removing a parameter.
Despite their higher computational cost, data-driven criteria generally lead to pruned networks with superior performance compared to data-free methods.

Data-driven pruning can be traced back to seminal works from the 1990s, including Optimal Brain Damage (OBD) \cite{lecun1990optimal} and Optimal Brain Surgery (OBS) \cite{Hssibi1992Surgeon}, which utilized second-order Taylor expansion to estimate the saliency of parameters for unstructured pruning. In OBD, the Hessian matrix is approximated as diagonal to reduce computational cost, with off-diagonal elements ignored. In this approach, overall saliency is computed by aggregating the isolated importance of individual elements, thereby neglecting their interconnections. However, as observed in OBS, the Hessian matrix is rarely diagonal, and ignoring parameter interdependencies can lead to inaccurate evaluation and significant degradation.  
Another key insight from OBS is that the remaining parameters must be recalibrated to achieve optimal pruning, a process that requires the participation of the nodes being pruned. Nevertheless, modern structural pruning techniques face limitations in these two aspects.  
First, existing criteria often follow the same diagonal paradigm (e.g., first-order Taylor \cite{molchanov2019importance} and second-order Fisher \cite{liu2021group}), which overlooks critical parameter interdependencies—a limitation identified in OBS that has yet to be fully addressed in modern networks.  
Second, naive pruning permanently discards parameters before the crucial fine-tuning stage, irretrievably losing their informational contribution and impeding the network's ability to recover.  
Although ``Soft Pruning'' methods \cite{asfp} allow pruned parameters to re-participate in ranking, they are ultimately discarded after a set number of fine-tuning epochs.

To address the aforementioned challenges, this paper proposes a structural pruning framework for a variety of architectures, termed \textbf{Optimal Brain Connection} (\textbf{OBC}). The framework consists of two key components:
\begin{enumerate}
\item \textbf{Jacobian Criterion} (\textbf{JC}), a computationally efficient yet highly accurate first-order metric, overcoming the off-diagonal effects and the inaccuracies of prior metrics. As illustrated  in \cref{fig:framework}(a), unlike Taylor \cite{molchanov2019importance} or Fisher-based Hessian \cite{liu2021group} criteria  that evaluate element-wise saliency in isolation, JC accounts for both intra-component (\eg, parameters within a filter) and inter-layer (\eg, a filter and its corresponding weight channels in the next layer) parameter connections, significantly reducing pruning-induced degradation.
\item \textbf{Equivalent Pruning} (\textbf{EP}),  a learnable transformation designed to maximize the informational capacity of pruned networks during fine-tuning, ensuring optimal parameter recalibration to approximate the original models. As illustrated in \cref{fig:framework}(b), EP employs a pair of transformation layers, $\tC$ and $\tD$, to respectively compress and decompress channels to the desired number, while retaining all original structural parameters during fine-tuning. After fine-tuning, as illustrated in \cref{fig:framework}(c), a permanent merge operation is conducted to obtain the same pruned model as naive pruning, but with improved performance.
\end{enumerate}

To demonstrate the effectiveness of OBC, we prune various models for computer vision, including Convolutional Neural Networks (CNNs) and Vision Transformers (ViTs). We also extend the task to object detection and natural language processing (NLP). 
Our ablation studies provide crucial insights into the effectiveness of OBC. First, disabling the interaction terms of JC leads to a significant performance drop, confirming that capturing parameter connections is the key driver of its superiority compared to the Taylor criterion. Second, the consistent performance gain from using EP validates our approach to capacity recovery during fine-tuning.

\section{Related Works}
\label{sec:Related Works}

\textbf{Model compression} can be broadly categorized into five types:
{{quantization}} \cite{lin2024awq},
{{low-rank approximation}} \cite{JSTSP},
{{neural architecture search}} \cite{wei2024auto},
{{knowledge distilling}} \cite{Yu_2025_CVPR},
and {{pruning}}~\cite{TPAMISurvey}. Pruning can further be divided into {{unstructured pruning}} and {{structural pruning}}, with this paper focusing on the latter.

\textbf{Unstructured pruning} zeros out weight elements, leading to irregular sparsity. The classical work by \citeauthor{han2015learning} (\citeyear{han2015learning}) iteratively pruned weights with magnitudes below a specified threshold, whereas a more efficient one-shot strategy was employed in subsequent research \cite{Wang2020Picking,mason2024makes}.
Unstructured pruning is also widely used in the Lottery Ticket Hypothesis \cite{LotteryTicket, gadhikar2024masks}, which posits that a highly sparse subnetwork performs comparably to the original dense network.

\textbf{Data-independent structural pruning} selects redundant components solely based on pretrained weight tensors.
Following the ``smaller-norm-less-important" assumption, norm-based methods \cite{PFEC} prune filters (or equivalently, channels) with the smallest norms, such as $\ell_1$ and $\ell_2$.
In particular,  BN Scale  \cite{BatchNormLiu2017ICCV} and  TLC \cite{liao2025till} leveraged statistics of layer or batch normalization (BN) layers to assess parameter importance. However, norm-based criteria become less effective when the variance of parameter norms is small. To address the limitation, relationship-based and hybrid criteria have been developed. Notable examples include WHC \cite{chen2023whc}, FPGM \cite{FPGM,kaparinos2025b}, CFP \cite{CFP} and GKP \cite{zhang2025flexible}. Instead of evaluating individual filters or channels, DepGraph \cite{fang2023depgraph} constructed dependency graphs for networks and removed coupled subnetworks with the lowest accumulated norms, achieving superior performance.
Most aforementioned methods employ either uniform pruning rates per layer (\ie, local pruning) or score normalization for global pruning to prevent excessive pruning in critical layers that may cause network collapse. To mitigate the issue, layer-wise sensitivity analysis was proposed \cite{PFEC}, while Isomorphic Pruning handled different component types separately \cite{fang2024isomorphic}. In contrast, our Jacobian Criterion achieves effective global pruning without normalization, owing to its accurate  saliency estimation.

\textbf{Data-driven structural pruning} either employs group-level regularization to enforce structural sparsity \cite{ding2021resrep,Guo_2025_CVPR,wang2020neural,huang2025pruning} or leverages data to assess the ``importance" of structural parameters \cite{Farina_2024_CVPR,GAL}. Various metrics exist to quantify importance, such as output reconstruction error \cite{NISP} and feature map rank \cite{HRank}. One of the most effective metrics is loss saliency \cite{NEURIPS2024_c1c44e46}, \ie, the degradation in empirical loss caused by the removal of a structural parameter. The smaller the saliency, the less important the structural parameters, making them safe for pruning. Loss saliency is often approximated using the second-order Taylor expansion \cite{nonnenmacher2022sosp}, where the first-order term is typically assumed to be zero \cite{liu2021group}.
As the Hessian matrix is computationally intensive, various techniques have been explored, including Fisher approximation \cite{11072282,mcgowan2024efficient,theis2018faster,liu2021group}, Hessian-vector products \cite{nonnenmacher2022sosp}, and Hessian-free approaches \cite{IFSO}.
Alternatively, some works adopt a first-order approximation \cite{you2019gate,PRE}, such as the popular Taylor criterion \cite{molchanov2019importance}. 
Note that Taylor and a great proportion of approximated second-order methods simply aggregate saliencies of individual weights or gating elements, while our proposed Jacobian Criterion accounts for parameter dependencies and intra-component interactions, enabling more accurate importance estimation.

\section{Methodology}
\label{sec:Methodology}

\subsection{Jacobian Criterion}
\label{sec:Jacobian Criterion}
Unlike the current best-performing first-order Taylor  \cite{molchanov2019importance} and the simplified second-order Fisher criterion \cite{liu2021group}, which assess parameters or gates in isolation, our approach captures both intra-component interactions and inter-layer dependencies. To achieve this, we evaluate the degradation of the empirical loss in a squared form when a perturbation $\Delta \w$ is applied to the vectorized weights $\w$ of a well-trained model:
\begin{align}
    L(\Delta\w)&\triangleq \sum_{n=1}^N \left[l(\x_n, \w+\Delta \w)-l(\x_n, \w)\right]^2 \nonumber\\
              &= \sum_{n=1}^N \left[l_n(\w+\Delta \w)-l_n(\w)\right]^2 \label{nnls} \\
               &=  \left[\l(\w+\Delta \w)-\l(\w)\right]^{\top}\left[\l(\w+\Delta \w)-\l(\w)\right] \nonumber
\end{align}
where $\x_n$ denotes the $n$-th sample batch, $l(\cdot)$ represents an arbitrary differentiable loss function (\eg, cross-entropy), $l_n(\w)\triangleq l(\x_n, \w)$  is the average loss for the $n$-th batch, and $\l\triangleq [l_1,\dots, l_N]^{\top}$.

\begin{figure}[t!]
\centering
\begin{subfigure}[t]{0.15\textwidth}
\captionsetup{justification=centering}
        \centering\includegraphics[width=\linewidth]{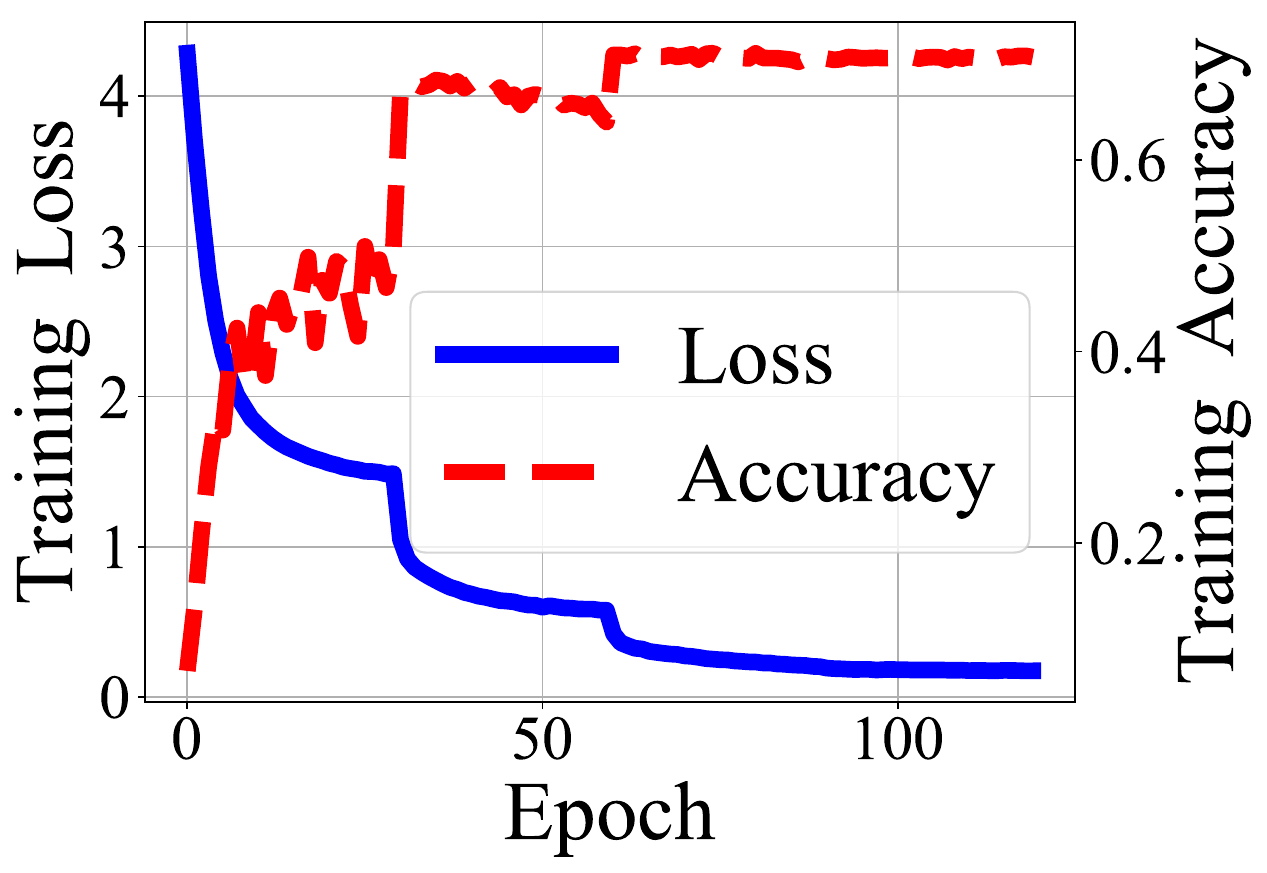}
\end{subfigure}
\begin{subfigure}[t]{0.15\textwidth}
\captionsetup{justification=centering}
        \centering\includegraphics[width=\linewidth]{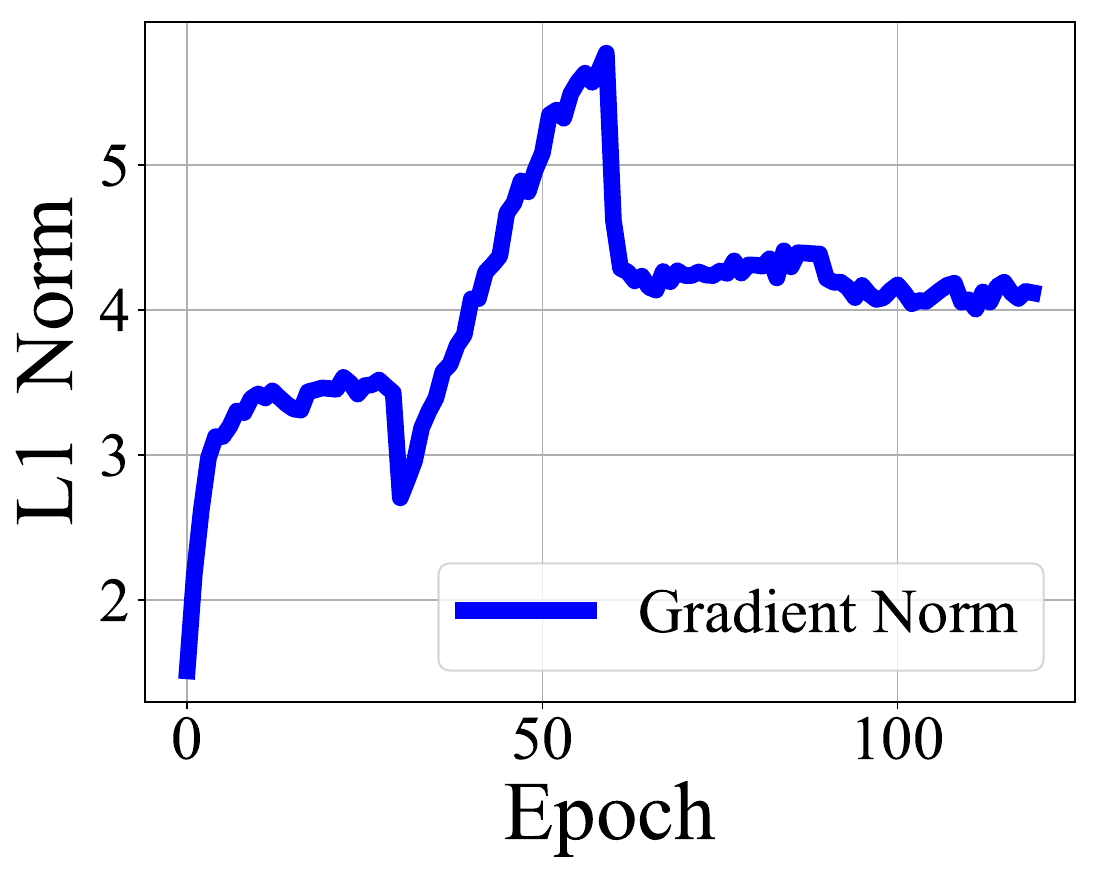}
\end{subfigure}
\begin{subfigure}[t]{0.15\textwidth}
\captionsetup{justification=centering}
        \centering\includegraphics[width=\linewidth]{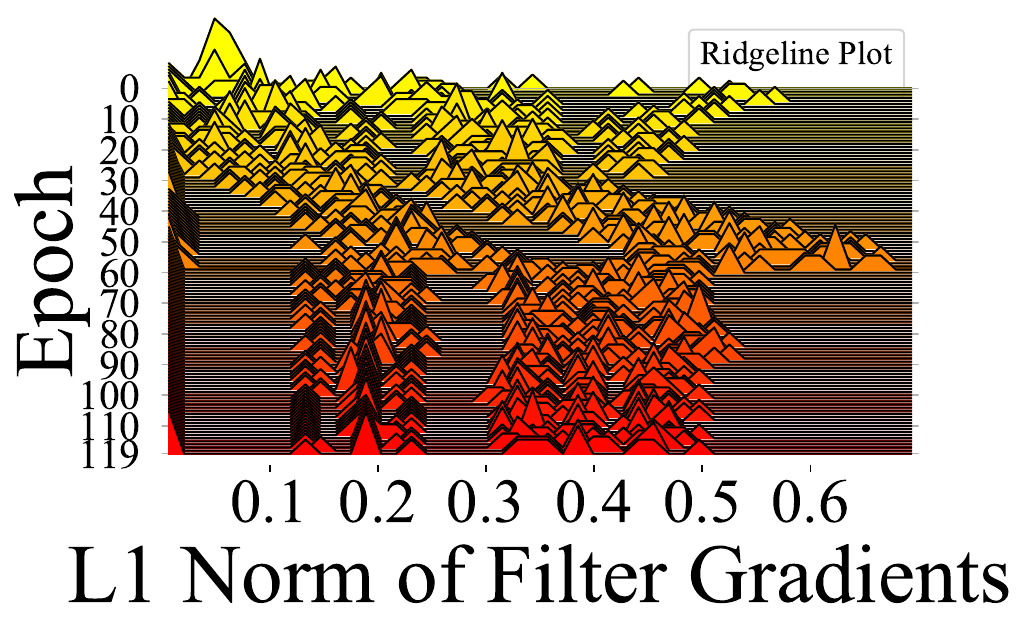}
\end{subfigure}
\caption{Gradients of deep neural networks do not necessarily converge to zero. {\bf Left}: Convergence process of ResNet-56 on CIFAR-100. {\bf Middle}: Gradient norm of Conv1 does not converge to zero during training. {\bf Right}: Gradient norms of filters in Conv1 vary significantly.
}
\label{fig:gradientNot0}
\end{figure}

When estimating the saliency of a converged network, 
second-order methods often assume that the gradient is zero and thus focus on estimating the second term of the Taylor expansion \cite{lecun1990optimal,liu2021group,frantar2022optimal}. 
However, as illustrated in \cref{fig:gradientNot0}, empirical evidence shows that the gradient does not necessarily converge to zero when the model stabilizes~\cite{zhang2022neural,chandramoorthy2022generalization}. Moreover, the gradient norms of structural parameters (such as filters) within the same layer can vary significantly.
Therefore, ignoring the first-order term is unreasonable. Given that the second-order term involving the Hessian matrix is computationally expensive in modern deep neural networks, we instead approximate the loss function vector $\l$ using only the first-order Taylor expansion:
\begin{equation}
    \l(\w+\Delta \w)\approx \l(\w)+\J\Delta \w+\mathcal{O}(\Vert\Delta \w\Vert),
\label{eq:firstTaylorEx}
\end{equation}
where $\J=\J(\w)$ denotes the Jacobian matrix, which consists of gradients 
$\J_{n,j}=\frac{\partial l_n(\w)}{\partial {w_j}}$.
Ignoring the error term and substituting \cref{eq:firstTaylorEx} into \cref{nnls}, we obtain
\begin{align}
    L(\Delta \w) =  {\Delta \w}^{\top} \J^{\top}\J\Delta \w.
    \label{Jacobian_full_JTJ}
\end{align}
Since $\J^{\top} \J$ is positive definite in practice, any nonzero perturbation in the weights will lead to a degradation of $L$.

\begin{figure}[t]
 \centering
    \includegraphics[width=0.46\textwidth]{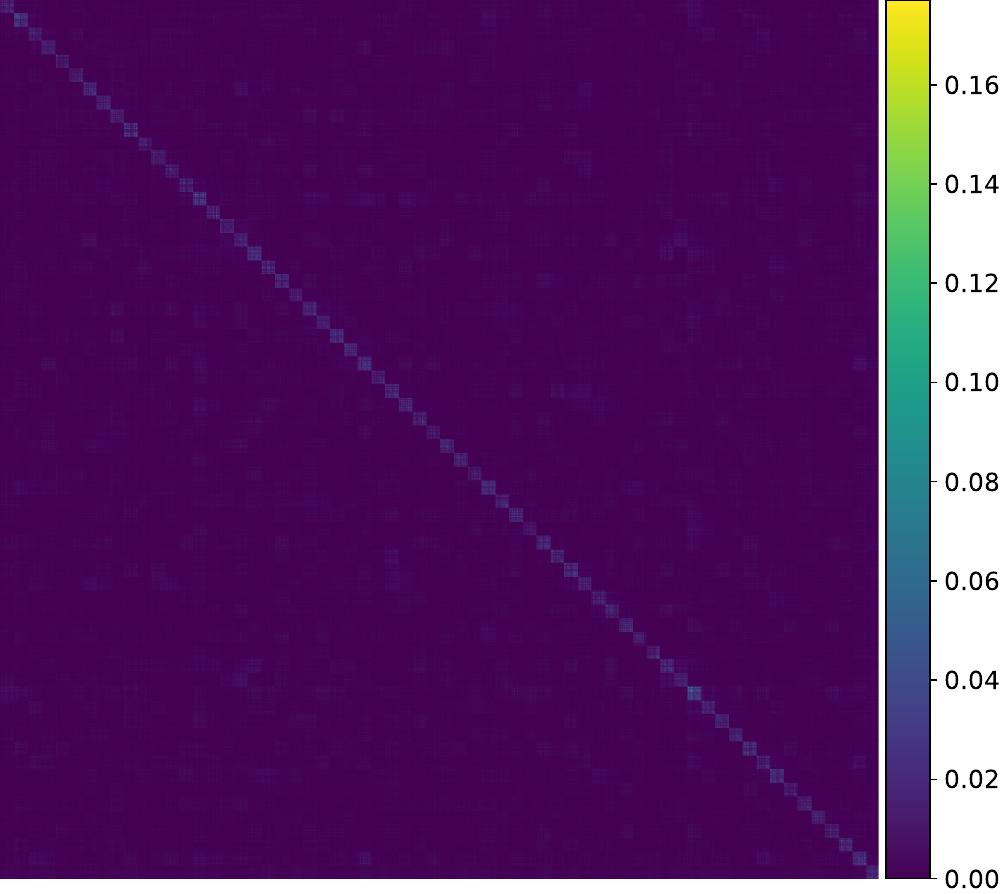}
    \caption{The diagonal blocks of $\mathrm{abs}(\J^{\top}\J)$ in the final convolutional layer of ResNet-56 are dominant. Note that here, $\J^{\top}\J$ \textbf{is block-diagonal, not diagonal}.  (Best viewed in color and zoom.)}
    \label{jtj_v}
\end{figure}

Calculating the full $\J^{\top} \J$ is computationally inefficient due to the vast number of parameters (ranging from millions to billions).  
To address the issue, we assume that only intra-component parameters (such as those within the same filter or channel) are correlated. Suppose there are $M$ structural parameters, then $\J^{\top}\J$ would be a block diagonal matrix, \ie, 
\begin{align}
\J^{\top}\J = 
\begin{pmatrix}
\J^{\top}_1\J_1 & \mathbf{0} & \cdots & \mathbf{0} \\
\mathbf{0} &\J^{\top}_2\J_2 & \cdots & \mathbf{0} \\
\vdots & \vdots & \ddots & \vdots \\
\mathbf{0} & \mathbf{0} & \cdots & \J^{\top}_M\J_M
\end{pmatrix}.
\end{align}
The assumption is not only necessary for efficient computation, but also aligns with the empirical practice that structural filters or channels within a layer are parallel and have limited direct interaction. (For example, the local $\J^{\top}\J$ of a full layer shown in \cref{jtj_v} is block-diagonal, instead of dense or diagonal.)
Under the assumption,  \cref{Jacobian_full_JTJ} would be 
\begin{align}
L(\Delta \w)&=\sum_{m=1}^M \Delta {\w}^\top_m\J_m^{\top}\J_m\Delta \w_m,\end{align}
where $\Delta \w=[\Delta {\w}^\top_1,\Delta \w_2^{\top},\cdots,\Delta \w_M^{\top}]^{\top}$.
Accordingly, we divide coupled or dependent structural parameters into $G$ groups with
\begin{equation}
\bigcup_{g=1}^G \mathbb{G}_g = \{1, \ldots, M\}, \quad \mathbb{G}_g \cap \mathbb{G}_{g'} = \emptyset \quad (\forall g \ne g')
\label{G_groups}
\end{equation}
and formulate the (one-step) structural pruning problem as selecting a group of coupled structural parameters, $\{\w_{m}|m\in \mathbb{G}_g\}$,
to minimize $L(\Delta \w)$:
\begin{equation}
\begin{aligned}
    \min_{g\in \{1,\cdots,G\}}&\quad \sum_{{m=1}}^{{M}} \Delta\w_{m}^{\top}\J_{m}^{\top}\J_{m}\Delta\w_{m} \\
\text{s.t.} \quad 
&\Delta \w_m =
\begin{cases}
-\w_m, & \text{if } m \in \mathbb{G}_g \\
0, & \text{otherwise}
\end{cases}
\end{aligned}
\end{equation}
that is 
\begin{align}
    \min_{g\in \{1,\cdots,G\}}  \mathcal{S}\left(\{\w_m|m\in\mathbb{G}_g\}\right)&\triangleq 
 \sum_{m\in \mathbb{G}_g}\mathcal{S}^{\rm (1)}(\w_m)\label{JacobianCriterionF}\\
 &\triangleq 
 \sum_{m\in \mathbb{G}_g} \w_m^{\top}\J_m^{\top}\J_m\w_m. \nonumber
\end{align}
We call $\mathcal{S}(\cdot)$ the {\bf Jacobian Criterion}, which quantifies the overall importance of the group by aggregating the individual saliency (defined as $\mathcal{S}^{(1)}(\cdot)$) of the coupled structural parameters in a summation form. For example, pruning $\w_m^{(l)}$, the $m$-th convolutional filter of the $l$-th layer, would also remove its downstream BN parameters $\b_{m}^{(l)} = [\gamma_m^{(l)}, \beta_m^{(l)}]^{\top}$ and the $m$-th input dimension of the $(l+1)$-th layer $\w_m'^{(l+1)}$. Thus, the saliency of the structural group is calculated by $
    {\mathcal{S}([\w_m^{(l)},\b_m^{(l)},\w_m'^{(l+1)}])}\nonumber={\mathcal{S}^{(1)}}\left(\w_ m^{(l)}\right)+{\mathcal{S}^{(1)}}\left(\b_m^{(l)}\right)+{\mathcal{S}^{(1)}}\left(\w_m'^{(l+1)}\right).$
A larger $\mathcal{S}$ indicates that the components are more important and should therefore be retained to prevent severe model degradation. Unlike previous works \cite{PRE,fang2023depgraph}, our Jacobian Criterion does not require normalization due to its accurate evaluation. In contrast, such normalization would significantly compromise the performance (see \cref{fig:ablationCriterion(b)}).

\begin{figure}[t]
 \centering
\includegraphics[width=0.46\textwidth]{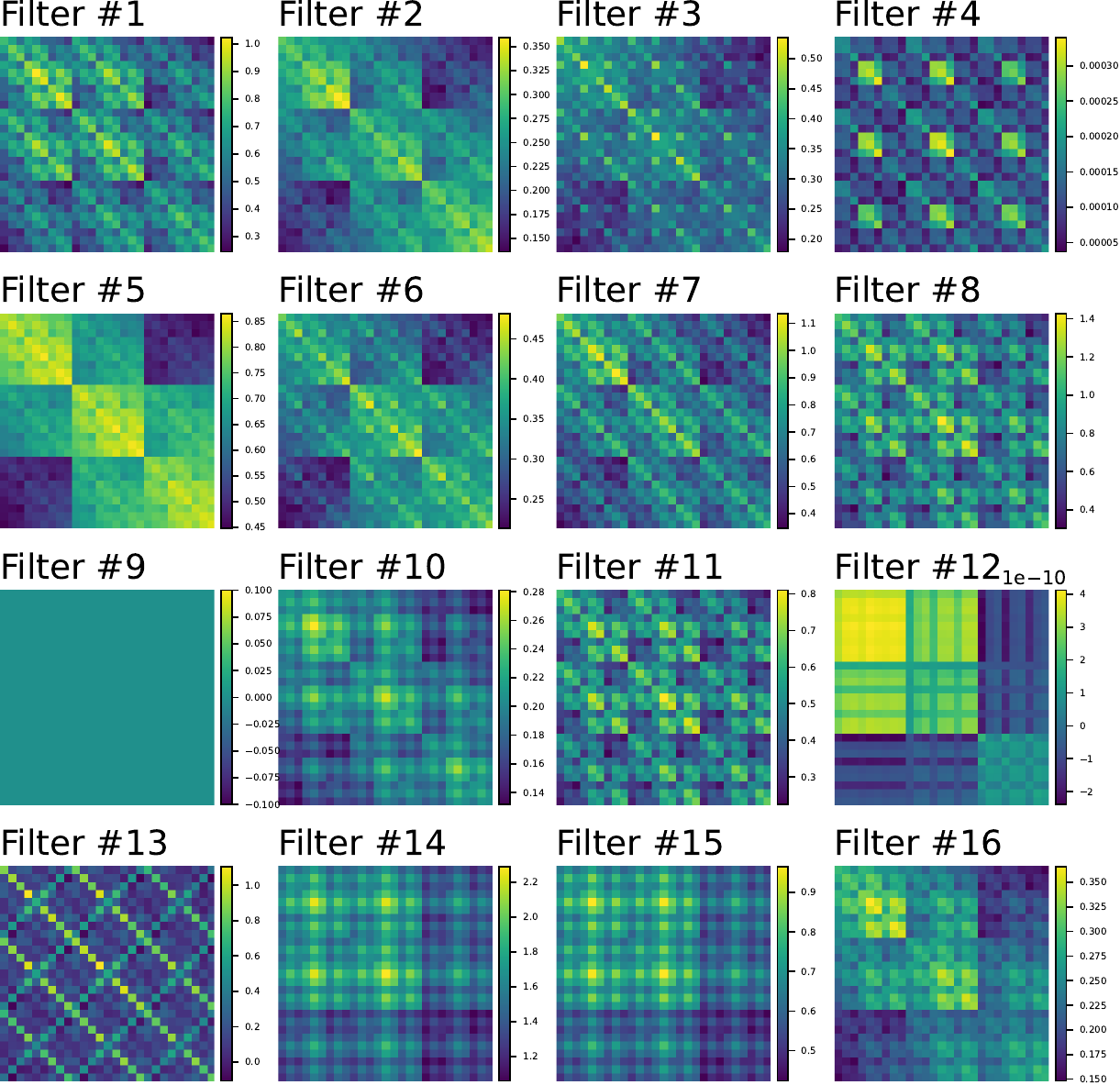}     
\caption{The $\J_m^{\top}\J_m$ of the 16 filters in Conv1 of ResNet-56 shows distinct patterns.
}
\label{fig:JTJ}
\end{figure}

\subsubsection{Discussion}
We reformulate several popular criteria and compare them with our JC (for a single structural weight) in \cref{tab:criteriaCompare}. JC can be seen as a generalization of the $\ell_2$ and Taylor criteria \cite{molchanov2019importance}, as well as a Gauss-Newton approximation of the dense second-order methods.
A key distinction of our JC is that it does not rely on the diagonal assumption but instead accounts for connections among structural parameters. As shown in \cref{fig:JTJ}, the off-diagonal elements of $\J_m^{\top}\J_m$ are non-negligible, highlighting strong parameter interactions within each structural parameter. Furthermore, $\J_m^{\top}\J_m$ for various structural filters exhibits distinct patterns.
Thus, compared to WHC \cite{chen2023whc}, which rescales $\Vert \w_m\Vert$ using a constant coefficient, leveraging $\J_m^\top\J_m$ for reweighting is more powerful in enhancing the discriminability.

\begin{table}[htbp]
\centering
\begin{tabular}{ll}
\toprule
Criterion & Formula \\
\midrule
$\ell_2$ norm &  $\Vert \w_m\Vert_2^2=\w_m^{\top}\I\w_m$ \\
Taylor    &  $\sum_i (w_ig_i)^2=\w_m^{\top}\left(\I\odot\J_m^{\top}\J_m\right)\w_m$ \\
WHC$^\dagger$ & $\left(\Vert \w_m \Vert_2\cdot \mho\right)^2=\w_m^{\top} ({\I\odot \mho^2}) \w_m$\\
Hessian*        &  $\sum_i w_i^2 h_{ii}=\w_m^{\top}\left(\I\odot\H_m\right)\w_m$ \\
\graycell \bf Jacobian (ours)  & \graycell $\w^{\top}_m\left(\J^{\top}_m\J_m\right)\w_m$  \\
\bottomrule
\end{tabular}
\raggedright\footnotesize $^\dagger$$\mho$ denotes the ``weighted dissimilarity"~\cite{chen2023whc}.  *$\H$ represents the Hessian matrix~\cite{liu2021group}.
\caption{Comparison of (individual-form) criteria: Our Jacobian Criterion takes into account the interconnections between parameters within the same structural weight. }
\label{tab:criteriaCompare}
\end{table}

\subsection{Equivalent Pruning}
\label{sec:Equivalent Pruning}
We propose a learnable transformation for ``softer pruning" that allows pruned parameters to participate in parameter recalibration during fine-tuning. Taking convolutional layers as an example (the MLP case can be easily extended), suppose there are two successive layers with weight tensors $\tW_1 \in \mathbb{R}^{O_1 \times I_1 \times K_1 \times K_1}$ and $\tW_2 \in \mathbb{R}^{O_2 \times O_1 \times K_2 \times K_2}$, where $O_i$, $I_i$, and $K_i$ represent the output, input, and kernel dimensions, respectively ($I_2 = O_1$). Instead of directly discarding the output channels of $\tW_1$ and input channels of $\tW_2$, we use paired linear layers, whose squeezed weights are $\C\in \mR^{\widehat{O}_1\times O_1}$ and $\D\in \mR^{\widehat{O}_1\times O_1}$, respectively, to reduce the number of output or input channels:
\begin{align}
    \widehat{\tW}_1 &\triangleq  \tW_1 \times_1 \C   \in \mR^{\widehat{O}_1\times I_1 \times K_1\times K_1}\label{eq:Compressor}\\
    \widehat{\tW}_2 &\triangleq  \tW_2 \times_2 \D  \in \mR^{{O}_2\times \widehat{O}_1 \times K_2\times K_2}\label{eq:Decompressor}
\end{align}
where ``$\times_n$" denotes the $n$-mode tensor multiplication \cite{kolda2009tensor}, and the pruned weight tensors are denoted as $\widehat{\tW}_1$ and $\widehat{\tW}_2$, with $\widehat{O}_1$ representing the remaining number of filters or channels ($\widehat{O}_1 < O_1$) after pruning.
Here $\C$ is the \textbf{Compressor} to reduce output dimension, and $\D$ is the corresponding 
\textbf{Decompressor}. Using $\C$ and $\D$, the network topology is equivalently modified to
\begin{align}
&\ \sigma\left(\tX_1 \otimes \widehat{\tW}_1\right) \otimes \widehat{\tW}_2 \\
=&\ \sigma\left(\tX_1 \otimes \left(\tW_1 \times_1 \C\right)\right) \otimes  \left(\tW_2\times_2 \D\right) \\
=&\ \sigma\left(\tX_1 \otimes {\tW}_1 \otimes \tC\right) \otimes \tD \otimes {\tW}_2
\label{CD_conv}
\end{align}
where $\sigma$ represents the non-linear operation, $\tX_1$ is the input tensor, $\otimes$ denotes the convolutional operation,   and 
\begin{align}
     \tC&={\rm unsqueeze(\C)}\in \mR^{\widehat{O}_1\times O_1\times 1 \times 1},\label{eq:tC}\\
  \tD&={\rm unsqueeze(\D^{\top})}\in \mR^{{O_1}\times \widehat{O}_1\times 1 \times 1}. \label{eq:tD}
\end{align}
We call this approach \textbf{Equivalent Pruning} (\textbf{EP}).
As shown in \cref{fig:framework}(b), EP imitates pruning by inserting two extra linear layers $\tC$ and $\tD$ before or after the original layers, forming  an autoencoder that performs feature fusion (dimension reduction) and re-mapping (dimension expansion). 

\textbf{Before fine-tuning}, with the redundant index set $\mathbb{P}$ recognized by our Jacobian Criterion in \cref{nnls}, we initialize
\begin{align}
    \C=\D=\I_{O_1}[\{1,2,\cdots,O_1\}\oslash \mathbb{P}, :].
\label{eq:init}
\end{align}
In this way,   
the initial output of the equivalently pruned model is identical to that of the naive pruning method.

\textbf{During fine-tuning}, as illustrated in \cref{fig:framework}(b), $\tC$ and $\tD$ form a learnable auto-encoder to implement feature merging and recovery. Notably, EP retains all original connections of $\tW_1$ and $\tW_2$, enabling the network to recalibrate its parameters by leveraging the full informational capacity.

\textbf{After finetuning}, as illustrated in \cref{fig:framework}(c), $\C$ and $\D$ can be merged into the original layers using \cref{eq:Compressor} and \cref{eq:Decompressor}, respectively, transforming the model from   \cref{fig:framework}(b) to \cref{fig:framework}(c). ({More implementation details can be found in the Appendix.}) This achieves the same pruned structure and inference Multiply–Accumulate Operations (MACs) as the naive pruning approach.  Unfortunately,  EP is not suitable for non-mergeable group convolution, which we leave as a problem for future study.

\begin{algorithm}[t!]
\footnotesize
\KwIn{Pretrained network $F$,  $N$ data batches,  target MACs $\tau$,  step pruning proportion $p$ ($<1$),  boolean $ep$.}
\KwOut{Pruned network $\widehat{F}$.}


{\color{gray}\textit{$\triangleright$ 0. Prepare}}

Divide parameters $\{\w_{m}\}_{m=1}^{M}$ into $G$ groups as \cref{G_groups}.
  
Initialize pruning indexes $\mathbb{P}=[\ ]$.
 
{\color{gray}\textit{$\triangleright$  1.  Jacobian Criterion Ranking (proposed)}}

\While{$\rm{MACs}>\tau$}{
    \For{$n=1 \to N$}{
         Obtain the $n$-th mini-batch gradients $\J[n,:]$.
    }
    \For{$g=1 \to G$}{
         Calculate saliency $\sum_{m\in \mathbb{G}_g}\w_{m}^{\top}\J_{m}^{\top}\J_{m}\w_{m}$.
    }
    Prune $(p\cdot G)$ groups with the lowest scores.
    
    Add the pruned indexes to $\mathbb{P}$.
}

{\color{gray}\textit{$\triangleright$ 2. Pruning and Finetuning}}

\If{\rm{not} \rm{$ep$}}{
{\color{gray}\textit{$\triangleright$  Naive Pruning}}

 Prune $F$ according to $\mathbb{P}$ to obtain the naively pruned $\widehat{F}$.
 
 Fine-tune  $\widehat{F}$.
}
\Else{

{\color{gray}\textit{$\triangleright$ Equivalent Pruning (proposed)}}

1.\ Initialize $\tC$ and $\tD$ via \cref{eq:tC}-\cref{eq:init} according to $\mathbb{P}$.
Insert $\tC$ and $\tD$ into the \textbf{unpruned} $F$ via \cref{CD_conv} to obtain the \textbf{equivalently pruned model} $\widehat{F}'$.

2.\ Fine-tune $\widehat{F}'$.
 
3.\ Merge $\tC$ and  $\tD$  into respective layers of $F$ via \cref{eq:Compressor} and \cref{eq:Decompressor}, to get \textbf{naively pruned model} $\widehat{F}$.
}
 
\KwRet Pruned network $\widehat{F}$.
\caption{One-shot Optimal Brain Connection}\label{alg:OBC}
\end{algorithm}

\subsection{Algorithm Description}
\cref{alg:OBC} shows our framework. Specifically, JC iteratively evaluates structural groups and prunes a small portion with the lowest score ($1/400$ for CIFAR and $1/100$ for ImageNet) until the desired pruning rate is achieved.
After pruning,  we perform fine-tuning to restore performance (with or without EP). While iteratively applying the ``pruning-finetuning" process could further improve performance \cite{liu2021group},  we opt for the one-shot manner for simplicity.

\section{Experiments}
\label{sec:Experiments}

\subsection{Settings}
\label{sec:Experimental Settings}
We evaluate our OBC on several widely used architectures for ImageNet and CIFAR, including CNN and Vision Transformer (ViT-B/16) \cite{DosovitskiyB2021ViT}. We also extend our evaluation to YOLOv7 for object detection and the large language model (LLM) ``Phi-3-mini-4k-instruct" for NLP. For all CIFAR experiments, we use a TITAN Xp GPU and repeat each experiment three times to report the average results. For ImageNet, experiments are conducted on four Nvidia 4090 GPUs. The implementation is based on DepGraph \cite{fang2023depgraph}, using Torch-Pruning v2.5.1. 

Unless otherwise specified, we set $N=50$ during pruning. The pruning step size is set to $p=0.25\%$ for CIFAR and $p=1\%$ for ImageNet. After pruning, models undergo one-shot fine-tuning with a small learning rate to restore performance. (See detailed settings in the Appendix.)

\begin{table}[t!]
  \centering
  \setlength{\tabcolsep}{1pt}
  \resizebox{\linewidth}{!}{
  \begin{tabular}{c l  c c  c}
      \toprule
      \multirow{2}{*}{\shortstack{\bf Model\\(MACs)}} & \multirow{2}{*}{\bf Method}   & \multirow{2}{*}{\shortstack{\bf Pruned\\\bf Acc. (\%)}} & \multirow{2}{*}{\shortstack{\bf $\Delta$ Acc.\\\bf (\%)}} & \bf \multirow{2}{*}{\shortstack{\bf MACs\\(B)}}  \\
      &&&&\\
    \hline  
      \multirow{11}{*}{\shortstack{ResNet-50\\\\4.13B}}
      & GReg-2~\cite{wang2020neural} & 75.36 & -0.77 & 2.77 \\ 
      & SOSP~\cite{nonnenmacher2022sosp} & 75.85 & -0.30  & 2.44 \\
      & SFP~\cite{asfp}   & 74.88 & -1.27 & 2.40 \\ 
      & FPGM~\cite{FPGM}   & 75.50 & -0.65 & 2.38 \\ 
      & Taylor~\cite{molchanov2019importance}  & 74.50 & -1.68 & 2.25 \\ 

      & Isomorph~\cite{fang2024isomorphic}  & 75.91 & -0.22 & 2.06 \\
      & Fisher~\cite{liu2021group}  & 76.42 & -0.37 & 2.04 \\ 
      & WHC \cite{chen2023whc} & 75.33 & -0.80 & \bf 1.92\\
      & DepGraph \cite{fang2023depgraph} & 75.83 & -0.32 &  1.99\\
      & \graycell \bf  Jacobian (ours) & \graycell  76.40 &\graycell  +0.25   & \graycell 2.03\\
      & \graycell \bf  Jacobian+EP (ours) & \graycell \bf 76.57 &\graycell \bf  +0.42      & \graycell 2.03\\
      \midrule

        \multirow{4}{*}{\shortstack{MobileNet-v2\\\\0.33B}} 
      & Meta~\cite{Meta}  & 68.20 & -6.50     & \bf 0.14  \\
      & Fisher~\cite{liu2021group} & 69.16 & -6.58   & 0.15 \\
      & DepGraph \cite{fang2023depgraph} & 68.46 & \bf-3.41    & 0.15 \\
      & \graycell \bf {Jacobian (ours)}  & \graycell 68.12 &  \graycell -3.75     & \graycell 0.15   \\
      \midrule

      \multirow{5}{*}{\shortstack{ViT-B/16\\17.59B}} 
      & AutoSculpt~\cite{jing2024autosculpt}  & 79.22 & -1.85 & 9.67 \\ 
      & CP-ViT~\cite{song2022cp}  & 76.75 & -1.16 & \bf 9.44 \\ 
       & DepGraph \cite{fang2023depgraph} & 79.58 & -1.39 & 10.40 \\
      & \graycell \bf Jacobian (ours) & \graycell  80.63  & \graycell -0.44    & \graycell  9.94 \\
       & \graycell \bf Jacobian+EP (ours) & \graycell \bf 80.85 & \graycell \bf -0.22 & \graycell  9.94 \\
      \bottomrule
      
  \end{tabular} 
  }
  \caption{Pruning results on ImageNet after fine-tuning.}
  \label{tbl:imagenet_pruning}
\end{table}

\begin{table}[t!]
  \centering
  \setlength{\tabcolsep}{1pt}
  \resizebox{\linewidth}{!}{
  \begin{tabular}{c l  l r  c}
      \toprule
      \multirow{2}{*}{\shortstack{\bf Model\\(MACs)}} & \multirow{2}{*}{\bf Method} &  \multirow{2}{*}{\shortstack{\bf  Pruned\\ \bf Acc. (\%)}} & \multirow{2}{*}{\shortstack{$\Delta$\bf  Acc.\\\bf (\%)}}  & \multirow{2}{*}{\shortstack{\bf MACs\\\bf Speedup}}\\
      &&&&\\
      \midrule  
        
        \multirow{11}{*}{\shortstack{VGG19\\\\CIFAR-100\\\\0.51B}} 
              & OBD \cite{wang2019eigendamage}    & 60.70 & -12.64   & 5.73$\times$   \\
              & OBD \cite{wang2019eigendamage}    & 60.66 & -12.68    & \bf 6.09$\times$  \\
              & \graycell \bf Jacobian (ours)                      & \graycell 71.68 & \graycell -1.82     &  \graycell 6.06$\times$  \\
              & \graycell \bf {Jacobian+EP (ours)}                   & \graycell \bf 72.27 & \graycell \bf -1.23     &\graycell 6.06$\times$  \\
              \cmidrule{2-5}
              & SOSP \cite{nonnenmacher2022sosp}  & 64.59 &-8.86  & 7.26$\times$ \\
              & EigenD \cite{wang2019eigendamage} & 65.18 & -8.16    & 8.80$\times$  \\
              & GReg-1 \cite{wang2020neural}      & 67.55 & -6.67& 8.84$\times$ \\
              & GReg-2 \cite{wang2020neural}      & 67.75 & -6.27  & 8.84$\times$ \\
              & DepGraph \cite{fang2023depgraph}    & 70.39 &  -3.11    & 8.92$\times$  \\
                & \graycell \bf {Jacobian (ours)}       & \graycell 70.41 &  \graycell -3.09  & \graycell \bf 8.98$\times$ \\
              & \graycell \bf {Jacobian+EP (ours)}                   & \graycell \bf 70.94 & \graycell \bf -2.56   & \graycell \bf 8.98$\times$ \\
      \midrule

      \multirow{15}{*}{\shortstack{ResNet-56\\\\CIFAR-10\\\\0.13B}} 
      & L1-norm~\cite{PFEC}                  & 91.80 & -1.00    & 2.00$\times$   \\
      & IFSO~\cite{IFSO}              & 93.65 & -0.03  & 2.00$\times$ \\
      & ASFP~\cite{asfp}                   & 93.12 & -0.47    & 2.11$\times$  \\
      & FPGM~\cite{FPGM}           & 93.26 & -0.33   & 2.11$\times$ \\
      & WHC \cite{chen2023whc}         & 93.66 & +0.07    & 2.11$\times$  \\
      & ResRep~\cite{ding2021resrep} & 93.71 & +0.00   & \bf 2.12$\times$ \\
      & DepGraph \cite{fang2023depgraph}     &  93.77 &  +0.24  &  2.11$\times$ \\ 
      & \graycell \bf {Jacobian (ours)}         & \graycell 93.83 & \graycell  +0.30   & \graycell  2.10$\times$  \\
      & \graycell \bf {Jacobian+EP (ours)}                       & \graycell \bf 93.92 & \graycell \bf +0.39    &  \graycell 2.10$\times$  \\
      
      \cmidrule{2-5}
      & GReg-1 \cite{wang2020neural}      & 93.18 & -0.18   & 2.55$\times$  \\
      & GReg-2 \cite{wang2020neural}      & 93.36 & -0.00    & 2.55$\times$  \\
      & WHC \cite{chen2023whc}             & 93.29 & -0.30      & \bf2.72$\times$  \\
      & DepGraph \cite{fang2023depgraph} &  93.64 &  +0.11     &  2.57$\times$ \\
      &  \graycell \bf {Jacobian (ours)}    & \graycell \bf 93.73  & \graycell \bf +0.20      & \graycell 2.51$\times$ \\
      & \graycell \bf {Jacobian+EP (ours)} & \graycell 93.71  & \graycell +0.18    &  \graycell 2.51$\times$\\     
      \bottomrule
  \end{tabular} 
    }
  \caption{Pruning results on CIFAR after fine-tuning. }
  \label{tbl:pruning_cifar}
\end{table}

\subsection{Pruning CNN and Transformer}  
\cref{tbl:imagenet_pruning,tbl:pruning_cifar} present the pruning results of OBC on the ImageNet and CIFAR  datasets (after fine-tuning), respectively. On ImageNet, OBC removes more than 50\% of the MACs in ResNet-50, not only maintaining accuracy but improving it by 0.42\%, which is 0.74\% higher than DepGraph.  On the CIFAR dataset, OBC achieves the highest accuracy at the maximum pruning rate for VGG19, demonstrating the precise estimation capability of the Jacobian Criterion. 
Furthermore, a comparison of results with and without EP shows that EP helps pruned networks adjust parameters during fine-tuning, mitigating performance degradation, especially for VGG.

\subsection{Ablation Study on the Jacobian Criterion}
\label{sec:Criteria}

\subsubsection{Apple-to-apple Comparison}

To validate the effectiveness of JC, we compare the raw performance degradation ({without fine-tuning}) of popular criteria under various pruning rates. The data-independent baselines include Random pruning, norm-based \textbf{Group L1} \cite{fang2023depgraph} and \textbf{BN Scale} \cite{BatchNormLiu2017ICCV}, relationship-based \textbf{FPGM} \cite{FPGM}, and the hybrid \textbf{WHC} \cite{chen2023whc}. The data-driven criteria include the first-order \textbf{Taylor} \cite{molchanov2019importance} and the second-order Fisher-based \textbf{Hessian} \cite{lecun1990optimal,liu2021group}. All criteria prune 1\%-5\% of structural parameters repeatedly per iteration until reaching a target threshold. \textbf{Note that for a fair comparison, all criteria adopt the same cross-layer importance aggregation as JC  (in summation form, see \cref{JacobianCriterionF}) without normalization in a global pruning manner.} Each experiment is repeated 5 times on CIFAR and 10 times on ImageNet. 

\begin{figure}[t!]
\centering
    \begin{subfigure}[t]{0.15\textwidth}
        \captionsetup{justification=centering}
        \centering\includegraphics[width=\linewidth]{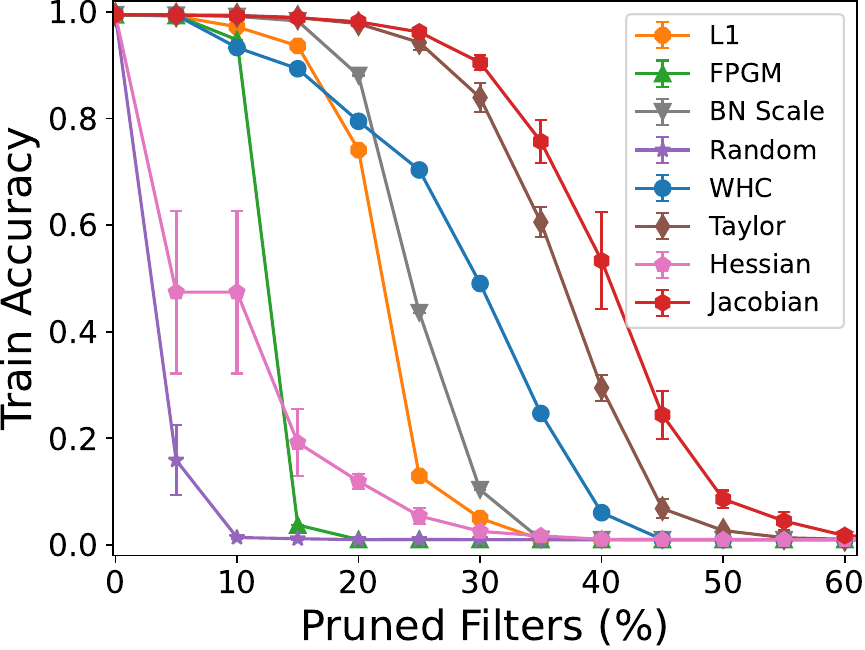}
        \caption{w.r.t. proportion}
        \label{fig:vggCriteria(a)}
    \end{subfigure}
    \hfill
    \begin{subfigure}[t]{0.15\textwidth}
        \captionsetup{justification=centering}
        \centering\includegraphics[width=\linewidth]
        {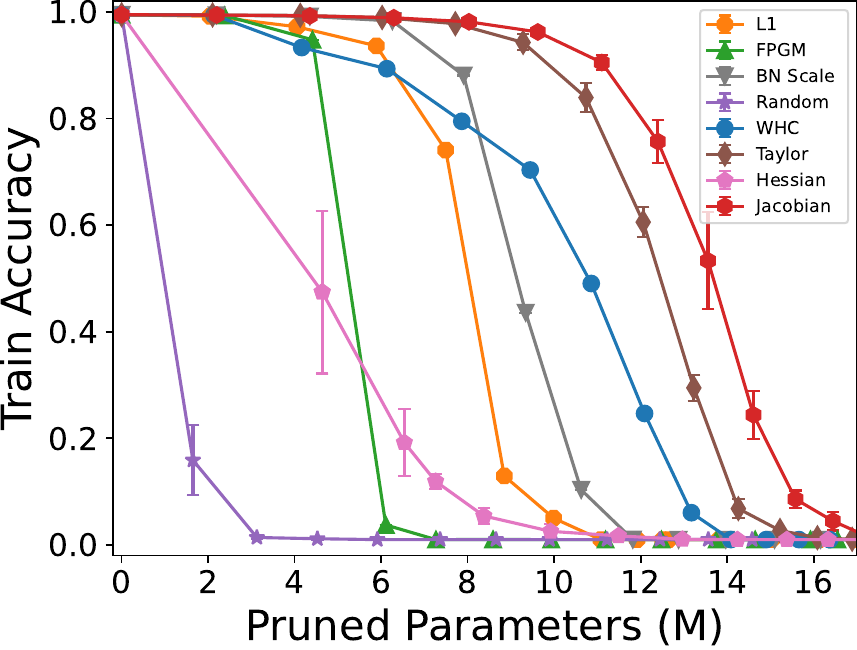}
        \caption{w.r.t. parameters}
        \label{fig:vggCriteria(b)}
    \end{subfigure}
    \hfill
    \begin{subfigure}[t]{0.15\textwidth}
        \captionsetup{justification=centering}
        \centering\includegraphics[width=\linewidth]{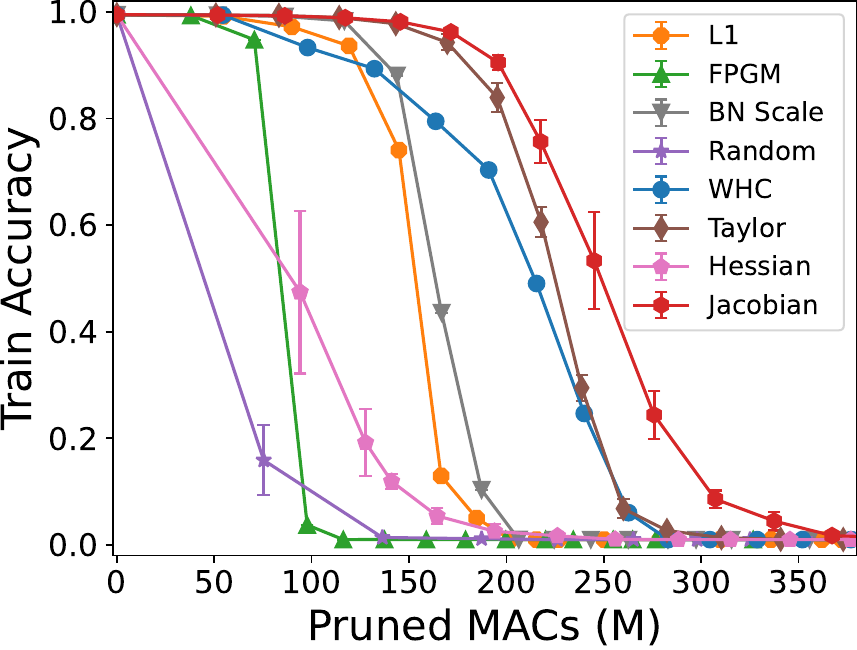}
        \caption{w.r.t. MACs}
        \label{fig:vggCriteria(c)}
    \end{subfigure}
    \caption{Pruned results  on VGG19 for CIFAR-100 \textbf{without fine-tuning}. Vertical lines represent standard deviations. 
    }
    \label{fig:vggCriteria}
\end{figure}

\begin{figure}[t!]
\centering
\begin{subfigure}[t]{0.23\textwidth}
        \includegraphics[width=\linewidth]{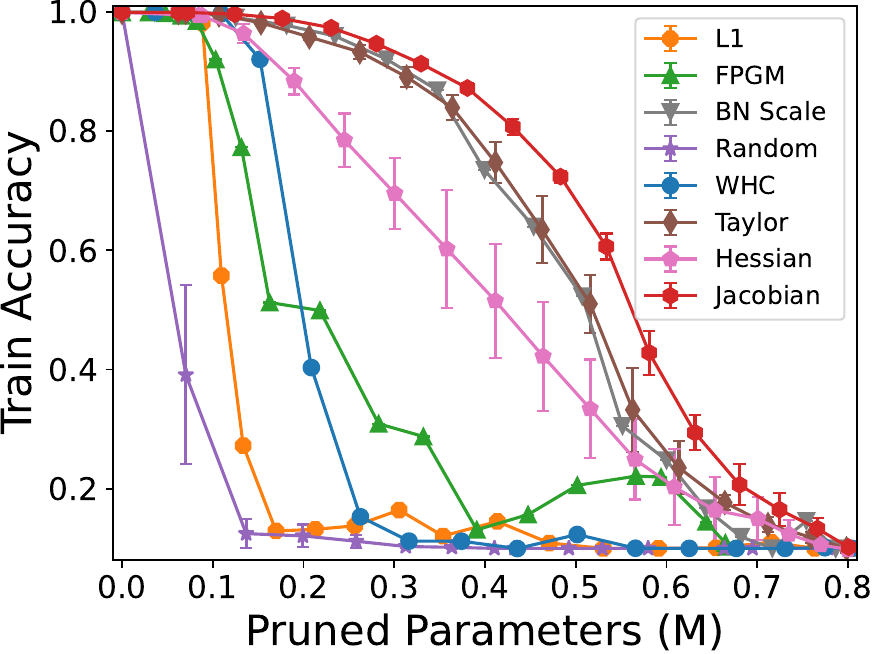}
        \caption{ResNet-56 for CIFAR-10}
        \label{fig:resnetCriteria(a)}
\end{subfigure}
\begin{subfigure}[t]{0.23\textwidth}
        \includegraphics[width=\linewidth]{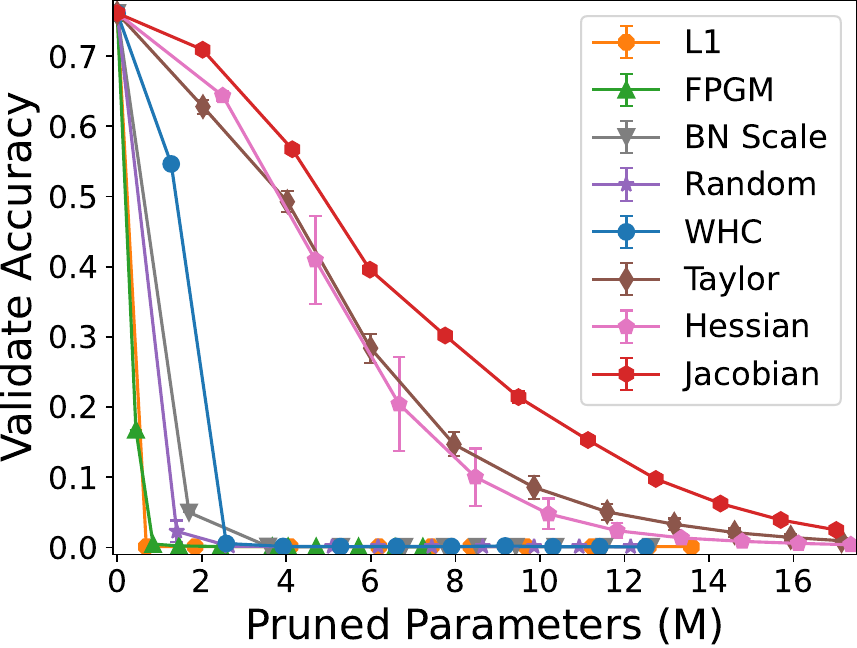}
        \caption{ResNet-50 for ImageNet}
        \label{fig:resnetCriteria(b)}
\end{subfigure}
\caption{Pruned results of various criteria on ResNet \textbf{without fine-tuning}. (a) $N=50$  (b)  $N=500$}
\label{fig:resnetCriteria}
\end{figure}

\begin{figure}[t!]
\centering
\begin{subfigure}[t]{0.23\textwidth}
\captionsetup{justification=centering}
        \centering\includegraphics[width=\linewidth]{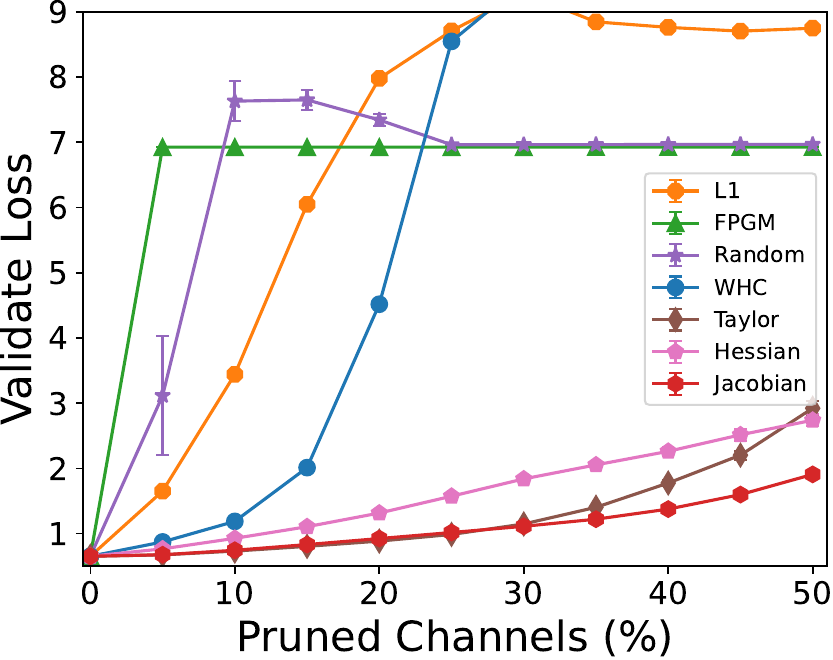}
        \caption{Global Pruning ($N=50$)}
        \label{fig:ViTCriteria(a)}
\end{subfigure}
\begin{subfigure}[t]{0.23\textwidth}
\captionsetup{justification=centering}
        \centering\includegraphics[width=\linewidth]{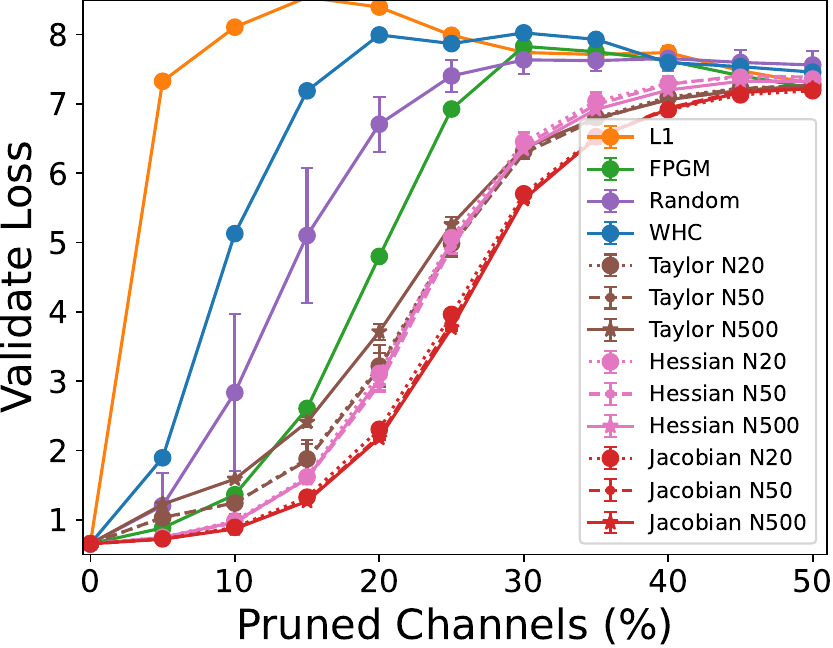}
        \caption{Uniform/Local Pruning}
        \label{fig:ViTCriteria(b)}
\end{subfigure}
\caption{Pruned results of various criteria on ViT-B/16 for ImageNet \textbf{without fine-tuning}.}
\label{fig:ViTCriteria}
\end{figure}

Results on \cref{fig:vggCriteria,fig:resnetCriteria,fig:ViTCriteria} show that JC consistently outperforms counterparts across datasets and models, resulting in minimal accuracy and loss degradation. The comparison between the Jacobian and Taylor criteria highlights the indispensable role of capturing interactions when evaluating the importance of structured parameters. Additionally, \cref{fig:ViTCriteria(a),fig:ViTCriteria(b)} show that global pruning significantly outperforms the local approach. 

\subsubsection{Time Complexity}

 \cref{tbl:criteriaTime} compares the average one-step evaluation time. While the Jacobian Criterion consumes marginally more computation time than data-independent methods, it remains comparable to the Taylor criterion and is significantly faster than the Hessian criterion.

\begin{table}[t]
    \centering
    \begin{tabular}{l r | l r}
        \toprule
        \multicolumn{2}{c|}{Data-Independent} & \multicolumn{2}{c}{Data-Driven} \\
        \cmidrule(lr){1-2} \cmidrule(lr){3-4}
        Criterion & Time (s) & Criterion & Time (s) \\
        \midrule
        Random &  \bf 0.07 & Taylor  &  \bf 2.66 \\
        BN Scale& 0.10 & \graycell \bf Jacobian (ours) & \graycell 2.73 \\
        Group L1 & 0.10 & Hessian& 242.80 \\
        FPGM & 0.15 & & \\
        WHC & 0.15 & & \\
        \bottomrule
    \end{tabular}
\caption{Average time consumption for per-step evaluation, tested on ResNet-56 for CIFAR-10 using a TITAN Xp GPU.} \label{tbl:criteriaTime}
\end{table}

\subsubsection{Parameter Interaction}
\label{sec:ablationstudy1_parameter}
\begin{figure}[h!]
\centering
\includegraphics[width=0.35\textwidth]{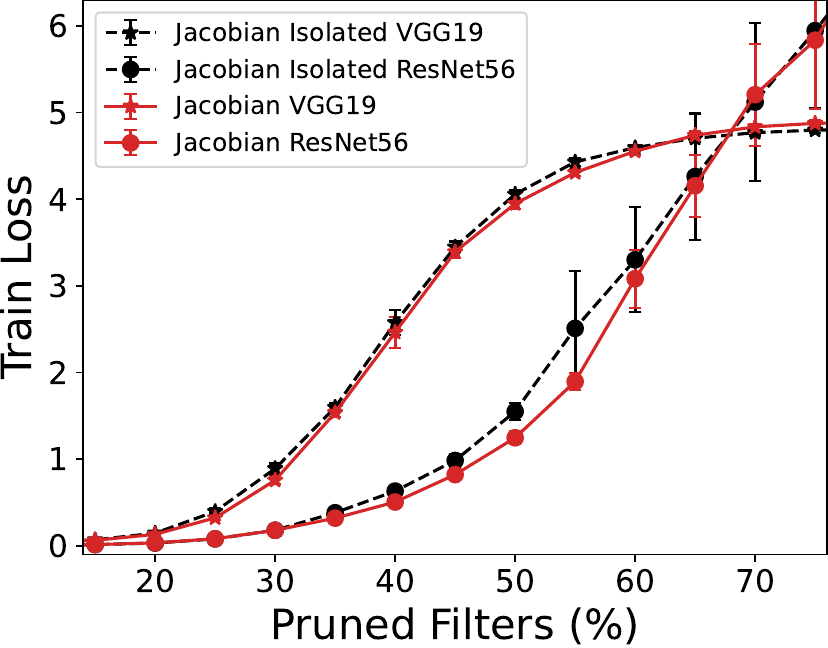}     
\caption{Diagonal vs. Dense $\J_m^\top\J_m$. Here the isolated version sets the $\J_m^\top\J_m$ of BN parameters to be diagonal.
}
\label{fig:ablationCriterion(a)}
\end{figure}

To assess the significance of parameter connections for importance estimation, we set non-diagonal elements of each $\J_m^\top\J_m$ for $\w_m=[\gamma,\beta]^\top$ in BN layers to zero, while retaining the full $\J_m^\top\J_m$ for other layers. As illustrated in \cref{fig:ablationCriterion(a)}, eliminating the cross-terms of $\J_m^\top\J_m$ on BN layers degrades pruning performance, while incorporating parameter connections in other layers still yields superior results to Taylor (see \cref{fig:vggCriteria(b)}). 
This demonstrates that parameter interactions modeled by a block-diagonal $\J^\top\J$ or dense $\J_m^\top\J_m$ are crucial for JC to assess saliency.

\subsubsection{Criterion Configurations}
\label{sec:ablationstudy1_settings}

\begin{figure}[t]
\centering
\includegraphics[width=0.35\textwidth]{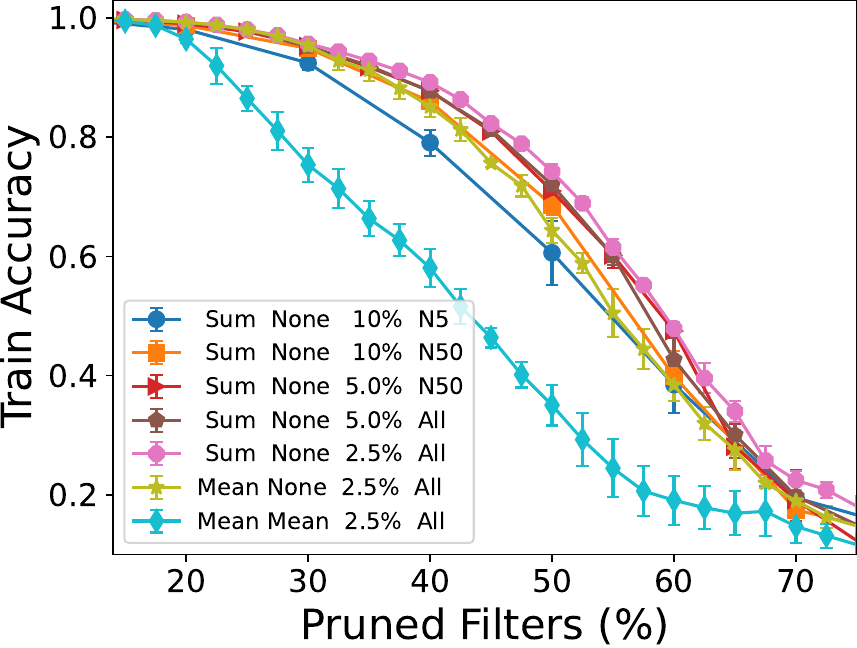}     
\caption{Ablation on JC's configurations. Here, ``Mean Mean 2.5\% All" represents that JC uses average aggregation,  per-layer mean normalization, $p=2.5\%$,  and all samples, and so on.}
\label{fig:ablationCriterion(b)}
\vspace{-0.5\baselineskip}
\end{figure}

We prune ResNet-56 on CIFAR-10 to study various configurations of JC, 
including 
aggregation strategies (summation ``Sum" vs. average ``Mean"), normalization methods (no normalization ``None" vs. per-layer mean normalization ``Mean"), pruning step sizes ($p \in \{10\%, 5\%, 2.5\%\}$), and sample sizes ($N \in \{5, 20, 50, 500\}$ and the option of using the full dataset ``All"). The results (see \cref{fig:ViTCriteria(b),fig:ablationCriterion(b)}) show that the default JC (``Sum" without normalization) effectively estimates saliency, while the ``Mean" aggregator and normalization reduce performance. For JC, $N=50$ performs similarly to $N=500$, with a slight improvement when using the full dataset, while Taylor suffers  significant estimation bias with $N=500$. Moreover, $p$ has a greater impact on evaluation quality than sample size, with smaller step sizes enabling better greedy optimality approximation.

\subsubsection{Task Extension} 
\begin{figure}[h]
    \centering
    \includegraphics[width=0.47\textwidth]{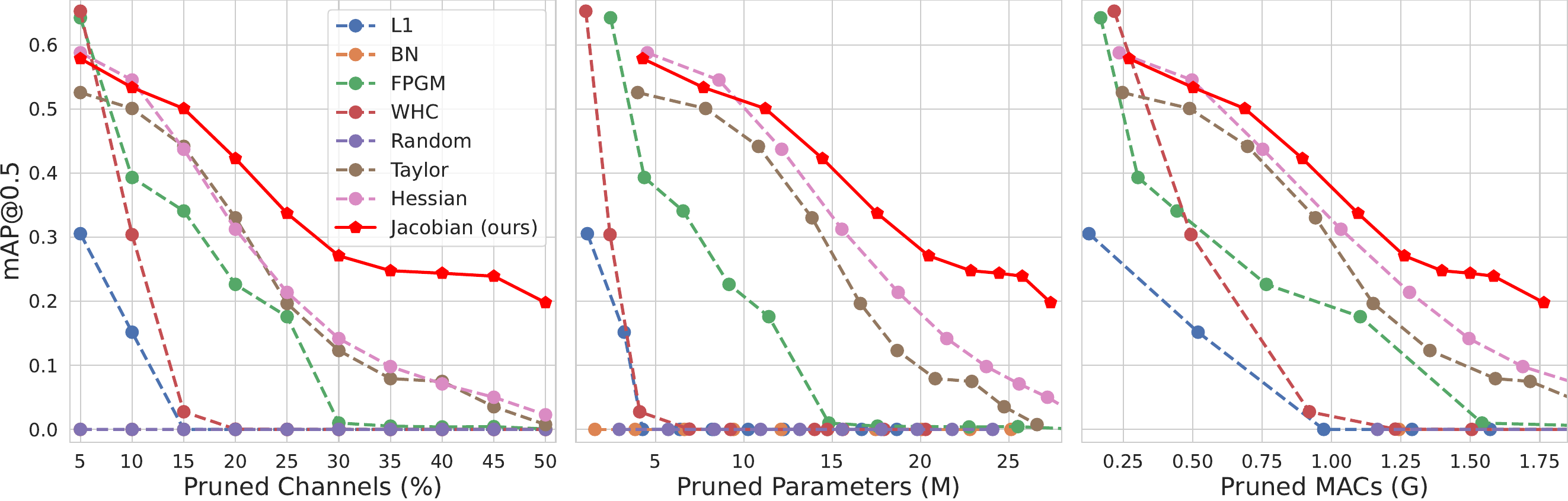}
    \caption{Pruned YOLOv7 \textbf{without fine-tuning}.}
    \label{yolo_figure}
\end{figure}

We explore the generality of JC on more complex models and tasks, including YOLOv7 for object detection and the LLM, Phi-3-mini-4k-instruct for NLP. As shown in \cref{yolo_figure} and \cref{LLM}, JC continues to outperform both data-free and data-driven criteria for vision and language models. Crucially, experiments reveal the stability of JC. In \cref{yolo_figure}, the YOLOv7 pruned by JC remains trainable, while the one pruned by Hessian is difficult to recover. In \cref{LLM}, the LLM pruned by group-norm collapses, while JC suffers the least degradation across all pruning ratios.

\begin{table}[h]
\centering
\small
\setlength{\tabcolsep}{1.5mm}
\begin{tabular}{l|cccc}
\toprule
Fine-tuning             & \multicolumn{2}{c}{\XSolidBrush} & \multicolumn{2}{c}{1 epoch} \\
\cmidrule(lr){1-1}               \cmidrule(lr){2-3} \cmidrule(lr){4-5}
 mAP@                 & 0.5       & 0.5:0.95      &0.5     & 0.5:0.95    \\
\midrule              
Hessian~\cite{liu2021group}            & 0.007         & 0.004             &  0.026           &    0.015             \\
\graycell \textbf{Jacobian (ours)} & \graycell \bf 0.193         & \graycell\bf  0.120             &  \graycell\bf 0.276       &\graycell \bf0.174        \\  
\bottomrule
\end{tabular}
\caption{mAP of YOLOv7 with 50\% of channels pruned.}
\label{yolotable}
\end{table}

\begin{table}[h!]
\centering
\small
\setlength{\tabcolsep}{1mm}
\resizebox{\linewidth}{!}{
\begin{tabular}{l|rrrrr}
\toprule
Pruned ratio*  & \textbf{5\%} & \textbf{10\%} & \textbf{15\%} & \textbf{20\%} & \textbf{25\%} \\
\midrule
Random           & 10.52         & 94.52        & 1216.49       & 4778.15      & 1.97E4        \\
Group $\ell_1$$^\dagger$\cite{fang2023depgraph}      & 3.07E6      & 2.78E6     & 7.33E5      & 4.07E5     & 5.78E5          \\
Hessian \cite{liu2021group}       & 8.00          & 31.90        & 188.95        & 777.91       & 2009.32            \\
\graycell \textbf{Jacobian (ours)}          &\graycell \textbf{7.66}          & \graycell\textbf{18.57}        & \graycell\textbf{101.62}        & \graycell\textbf{476.42 }      & \graycell\textbf{1131.52}         \\
\bottomrule
\end{tabular}
}
\raggedright\footnotesize *Original perplexity: 5.63.  $^\dagger$The model collapses.  
\caption{The perplexity of the pruned LLM Phi-3-mini-4k-instruct {\bf without fine-tuning} evaluated using WikiText-2. }
\label{LLM}
\end{table}

\begin{table*}[t!]
\centering
\begin{tabular}{llccccccc}
\toprule

\multirow{3}{*}{Normalizer} & \multirow{3}{*}{Aggregator} & \multirow{3}{*}{Method} & \multirow{3}{*}{\shortstack{\bf  EP\\\bf (ours)}} & \multicolumn{5}{c}{MACs Speedup} \\

\cmidrule(lr){5-9}
 &  &  &  & 1.5$\times$  & 3$\times$ & 6$\times$ & 9$\times$ & 12$\times$ \\

\midrule

\multirow{16}{*}{\bf{None}} & \multirow{5}{*}{Max} & \multirow{2}{*}{DepGraph } & \XSolidBrush & 73.22   & 73.29   & 71.75   & 69.86   & \bf 64.66   \\
 &  &  & \Checkmark & \bf 73.67   & \bf 73.36   & 71.75   & \bf 69.96   & 64.16   \\

 \cmidrule(lr){3-9}
 &  &  \multirow{2}{*}{\bf \shortstack{Jacobian\\(ours)}   } & \XSolidBrush & 73.27   & 73.22   & 71.86   & 70.05   & 68.56   \\
 &  &  & \Checkmark & \bf 73.46   & \bf 73.58   & \bf 72.35   & \bf 70.64   & \bf 69.05   \\

 \cmidrule{2-9}\cmidrule{2-9}
 & \multirow{5}{*}{Mean} & \multirow{2}{*}{DepGraph } & \XSolidBrush & 73.31   & 73.28   & 71.26   & \bf 68.17   & \bf 58.03   \\
 &  &  & \Checkmark & \bf 73.78   & \bf 73.43   & \bf 71.78   & 67.99   & 57.46   \\

 \cmidrule(lr){3-9}
 &  &  \multirow{2}{*}{\bf \shortstack{Jacobian\\(ours)}   } & \XSolidBrush & 73.56   & 73.22   & 71.93   & 70.29   & 68.15   \\
 &  &  & \Checkmark & \bf 73.86   & \bf 73.55   & \bf 71.99   & \bf 70.78   & \bf 68.23   \\

  \cmidrule{2-9}\cmidrule{2-9}
 & \multirow{5}{*}{\bf\shortstack{Sum}} & \multirow{2}{*}{DepGraph } & \XSolidBrush & 73.30   & \bf 73.34   & 71.04   & 67.63   & 56.55   \\
 &  &  & \Checkmark & \bf 73.67   & 73.30   & \bf 72.02   & \bf 68.20   & \bf 57.89   \\

 \cmidrule(lr){3-9}
 &  &  \multirow{2}{*}{\bf \shortstack{Jacobian\\(ours)}   } & \XSolidBrush & 73.66   & 73.09   & 71.68   & 70.41   & 68.14   \\
 &  &  & \Checkmark & \bf 73.84   & \bf 73.53   & \bf 72.27   & \bf 70.94   & \bf 68.61   \\

\midrule

\multirow{5}{*}{Mean} & \multirow{5}{*}{Mean} & \multirow{2}{*}{DepGraph } & \XSolidBrush & 73.17   & 73.25   & 71.47   & 68.40   & 66.99   \\
 &  &  & \Checkmark & \bf 73.75   & \bf 73.40   & \bf 71.99   & \bf 68.44   & \bf 67.00   \\

 \cmidrule(lr){3-9}
 &  &  \multirow{2}{*}{\bf \shortstack{Jacobian\\(ours)}   } & \XSolidBrush & 73.25   & 73.22   & 35.94   & 28.11   & 51.69   \\
 &  &  & \Checkmark & \bf 73.74   & \bf 73.64   & \bf 38.08   & \bf 31.18   & \bf 53.58   \\
 \bottomrule
\end{tabular}
\caption{Fine-tuned accuracy (\%) of pruned VGG-19 on CIFAR-100 under various MACs, with or without EP.}

\label{tbl:EPVgg19Ablation}
\end{table*}

\subsection{Ablation Study on EP and Settings}
\label{sec:Evaluating Equivalent Pruning}
We prune VGG19 on CIFAR-100 under various settings, including different importance normalizers (``None" and ``Mean"), aggregators (``Max", ``Mean", and ``Sum"), and MACs speedup factors.  
We also reimplement DepGraph~\cite{fang2023depgraph} with and without EP to assess EP's generalizability.  
The results in \cref{tbl:EPVgg19Ablation} demonstrate that EP generally enhances the accuracy of pruned models after fine-tuning across all pruning rates and criterion settings, achieving a maximum improvement of over $1\%$ for both OBC and DepGraph. This highlights the importance of retaining the full informational capacity of the original model for optimal parameter recalibration during fine-tuning.

\section{Conclusion}
Inspired by OBS \cite{Hssibi1992Surgeon}, this paper revisits parameter interactions in structural pruning and proposes the Optimal Brain Connection (OBC) framework, which comprises two components: the Jacobian Criterion and the Equivalent Pruning mechanism.  
To identify redundancy, OBC formulates structural pruning as a least squares problem that minimizes the squared loss perturbation induced by pruning, from which the simple yet effective Jacobian Criterion is derived. This criterion captures both intra-component interactions and inter-layer dependencies, enabling accurate estimation of structural parameter saliency.  
To further enhance fine-tuned performance, Equivalent Pruning employs paired compressor-decompressor autoencoders, ensuring all original structural parameters contribute to network recalibration.  
Extensive experiments on both CNNs and Transformers demonstrate OBC's effectiveness in redundancy elimination and performance preservation.

\bibliography{cite}

\onecolumn
\clearpage
\section{Appendix}
\section{Limitations and Future Work}
The compressors and decompressors of Equivalent Pruning (EP) introduce additional computational overhead during fine-tuning. However, these added modules are lightweight $1\times1$ convolutional or linear layers, which scale efficiently for large models. EP merges layers post-fine-tuning, resulting in  a pruned model identical to a naively pruned model, yet with significantly improved performance. Given the resource constraints for deployment and the abundant resources available for fine-tuning, we argue that the trade-off in fine-tuning is justifiable.

While EP proves highly effective across a broad range of standard CNNs and Transformers, the current merging strategy is not directly compatible with group convolutions. We leave this limitation for future work.

\section{Hyper-parameters}
\label{sec:hyperSettings}
\cref{tbl:hyper} summarizes the hyperparameters used for fine-tuning.

\begin{table*}[h!]
\centering

\begin{tabular}{lccccc}
\toprule
\multirow{2}{*}{Model}            & \multicolumn{2}{c}{CIFAR}   & \multicolumn{3}{c}{ImageNet}                        \\
\cmidrule(lr){2-3}  \cmidrule(lr){4-6}
  & VGG19        & ResNet-56    & ResNet-50    & MobileNet-v2         & ViT-B/16      \\
\midrule
sparse learning \cite{fang2023depgraph}             & \Checkmark   & \Checkmark   & \Checkmark   & \Checkmark                & \XSolidBrush  \\
$N$                                 & 50           & 50           & 50           &   50                   & 50            \\
Jacobian batch size                 & 128          & 128          & 256          & 256 & 64            \\
$p$                                 & 1/400        & 1/400        & 1/100        &  1/100                    & 1/100         \\
optimizer                           & SGD          & SGD          & SGD          & SGD                  & AdamW         \\
base learning rate (lr)                & 0.01         & 0.01         & 0.04         & 0.036                     & 0.000125      \\
base lr for $\tC$ and  $\tD$ & 0.002        & 0.02         & 0.01         & - & 0.000025      \\
learning rate schedule              & 60, 80       & 60, 80       & cosine       & cosine               & cosine        \\
weight decay (wd)                        & 0.0005       & 0.0005       & 0.001        &  0.00002                    & 0.05          \\
wd for  $\tC$ and $\tD$    & 0.0005       & 0            & 0            & -             \\
optimizer momentum                  & 0.9          & 0.9          & 0.9          & 0.9                  & (0.9, 0.999)  \\
batch size                          & 128          & 128          & 256*4        &   512*4                   & 128*4         \\
total tuning epochs               & 100          & 100          & 100          &  300                    & 100           \\
warmup epochs                       & 0            & 0            & 0            & 0                    & 30            \\
warmup decay                & -         & -         & -         & -                 & linear, 0.033 \\
distillation coefficient            & -           & -            & 0.5          & -                    & 0.5           \\
distillation T                      & -            & -            & 4            & -                    & 4             \\
mixup                               & \XSolidBrush & \XSolidBrush & \XSolidBrush & \XSolidBrush         & 0.8           \\
cutmix                              & \XSolidBrush & \XSolidBrush & \XSolidBrush & \XSolidBrush         & 1.0           \\
random erasing                      & \XSolidBrush & \XSolidBrush & \XSolidBrush & \XSolidBrush         & 0.25          \\
label smoothing                     & \XSolidBrush & \XSolidBrush & \XSolidBrush & \XSolidBrush         & 0.11          \\
gradient clip                       & \XSolidBrush & \XSolidBrush & \XSolidBrush & \XSolidBrush         & 1             \\
exp. mov. avg. (EMA)                & \XSolidBrush & \XSolidBrush & \XSolidBrush & \XSolidBrush         & 0.99998  \\
auto mixed precision (AMP)       & \XSolidBrush & \XSolidBrush & \Checkmark & \Checkmark         & \Checkmark  \\
  \bottomrule
\end{tabular}
\caption{Details of hyperparameters for fine-tuning.}
\label{tbl:hyper}
\end{table*}

\section{Pruning Implementation Details}
\label{details_imple}

\begin{itemize}
    \item For MobileNet-v2, we do not prune the last convolutional layer. Since EP is not suitable for non-mergeable group convolutions, it is not applied to MobileNet.
    \item For ViT, only the feed-forward modules are pruned. The original structure is transformed by EP from 
    \begin{equation}
        \text{Linear - GELU - Linear} \nonumber
    \end{equation}
    to 
    \begin{equation}
    \text{Linear - }\tC\text{ - GELU - }\tD\text{ - Linear}. \nonumber
    \end{equation}
    
    \item For ResNet on ImageNet, shortcuts are not pruned. For VGG and ResNet on CIFAR, all layers are pruned. Their original structure is transformed by EP from 
    \begin{equation}
        \text{Conv - BN - ReLU - Conv/Classifier} \nonumber
    \end{equation}
    to 
    \begin{equation}
    \text{Conv - }\tC\text{ - BN - ReLU - }\tD\text{ - Conv/Classifier}. \nonumber
    \end{equation}
\end{itemize}

Furthermore, we find that for ResNet, the structure ``Conv-BN-$\tC$-New\_BN-ReLU-$\tD$-Conv/Classifier", where the ``New\_BN" is identically initialized before fine-tuning, yields a slight performance improvement of about $0.2\%$. However, this approach performs  worse on VGG. Therefore, for simplicity, we exclude it from our main method.

\end{document}